\documentclass[runningheads]{llncs}

\usepackage{graphicx}
\usepackage{booktabs}

\usepackage[accsupp]{axessibility}  

\usepackage{hyperref}

\usepackage{orcidlink}

\usepackage{times}
\usepackage{graphicx}
\DeclareGraphicsExtensions{.pdf,.png,.jpg}
\usepackage{amsmath}
\usepackage{amssymb}  
\usepackage{pifont}
\newcommand{\xmark}{\ding{55}}%
\usepackage{makecell}
\usepackage{booktabs}
\usepackage{multirow}
\usepackage{here}
\usepackage{kotex}
\usepackage{soul, color}
\usepackage{cuted}
\usepackage{array}
\usepackage{threeparttable}
\usepackage{caption}
\usepackage{cleveref}
\usepackage[table]{xcolor}
\definecolor{bestcolor}{HTML}{bce6cd}
\definecolor{secondcolor}{HTML}{e4eebc}
\definecolor{thirdcolor}{HTML}{fef8c4}
\definecolor{BestGreen}{HTML}{2F9E44} 
\newcommand{\bestbg}[1]{\cellcolor{bestcolor}\textbf{#1}}
\newcommand{\secondbg}[1]{\cellcolor{secondcolor}#1}
\newcommand{\thirdbg}[1]{\cellcolor{thirdcolor}#1}

\begin{document}

\title{LangGS-SLAM: Real-Time Language-Feature Gaussian Splatting SLAM}

\author{Seongbo Ha\orcidlink{0009-0007-7018-1598} \and
Sibaek Lee\orcidlink{0009-0007-6600-2323} \and
Kyungsu Kang\orcidlink{0000-0002-6366-9735} \and
Joonyeol Choi\orcidlink{0009-0002-2276-9876} \and
Seungjun Tak\orcidlink{0009-0008-2204-6137} \and
Hyeonwoo Yu\orcidlink{0000-0002-9505-7581}}
\authorrunning{Seongbo Ha et al.}

\institute{Sungkyunkwan University, Suwon, South Korea\\
}
\maketitle


\begin{figure}[t]
  \centering
   \includegraphics[width=\linewidth]{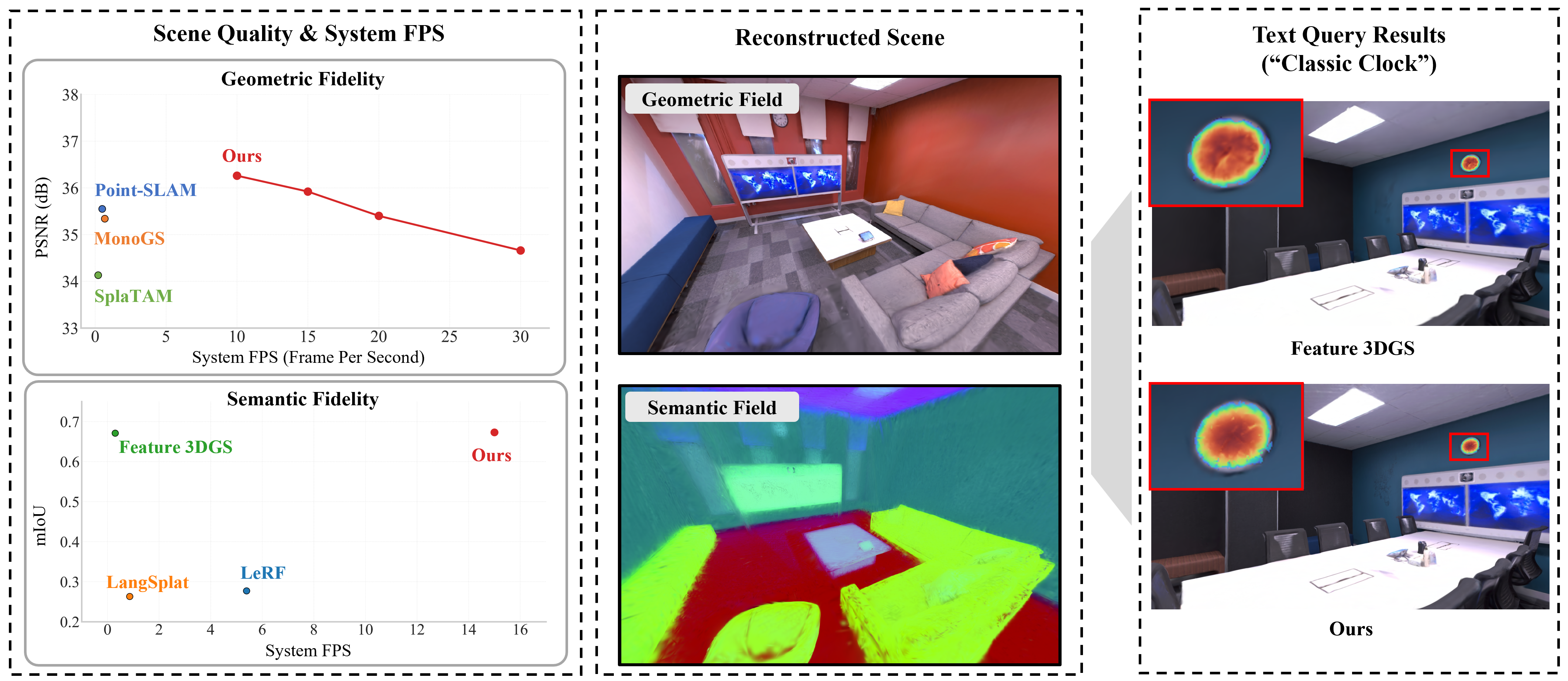}
   \caption{
   We construct a language-feature aligned 3DGS field online from RGB-D input.
   The reconstructed semantic–geometric map supports text-driven 3D queries for interactive perception.
   Despite reconstructing complex semantic–geometric scenes, our method surpasses geometric-only SOTA in geometric fidelity and matches offline dense VLM methods in semantic quality, while running 50× faster.
   }
   \label{fig:title}
   \vspace{-15pt}
\end{figure}

\begin{abstract}
In this paper, we propose a RGBD SLAM system that reconstructs a language-aligned dense feature field while sustaining low-latency tracking and mapping.
First, we introduce a Top-K Rendering pipeline, a high-throughput and semantic-distortion-free method for efficiently rendering high-dimensional feature maps.
To address the resulting semantic–geometric discrepancy and mitigate the memory consumption, we further design a multi-criteria map management strategy that prunes redundant or inconsistent Gaussians while preserving scene integrity.
Finally, a hybrid field optimization framework jointly refines the geometric and semantic fields under real-time constraints by decoupling their optimization frequencies according to field characteristics.
The proposed system achieves superior geometric fidelity compared to geometric-only baselines and comparable semantic fidelity to offline approaches while operating at 15 FPS.
Our results demonstrate that online SLAM with dense, uncompressed language-aligned feature fields is both feasible and effective, bridging the gap between 3D perception and language-based reasoning.
%
\end{abstract}
\section{Introduction}
\label{sec:intro}

Large Language Models (LLMs) are extending their capabilities beyond pure language understanding to act as reasoning engines for embodied AI systems~\cite{palm-e,saycan,rt2}.
Recent works such as ConceptFusion~\cite{conceptfusion}, LERF~\cite{kerr2023lerf}, CLIP-Fields~\cite{clip-fields}, and 3D-LLM~\cite{3d-llm} demonstrate that language-aligned 3D feature fields enable open-vocabulary queries and spatial reasoning, allowing LLMs to interpret and interact with 3D environments.

However, most existing semantic SLAM~\cite{sni-slam,li2024sgs,zhu2024semgauss,li2024dns} rely on closed-set semantic labels or scene-specific features.
Such representations limit open-vocabulary reasoning and prevent direct interaction with LLMs.
In contrast, Vision-Language Models (VLMs)~\cite{clip,lseg,openseg} offer continuous, language-aligned embeddings that encode rich and generalizable semantics.
Preserving these embeddings in 3D would enable open-set segmentation and natural-language reasoning directly over spatial representations—an essential step toward LLM-interactive perception.

To this end, we argue that a dense language-aligned feature field SLAM that directly aligns language features in 3D space while maintaining geometric consistency.
Recent works~\cite{feature-3dgs,langsplat} have shown the feasibility of constructing such fields offline, and Online Language Splatting~\cite{onlinelsplatting} extended this idea to online reconstruction.
However, its system speed remains under 1 FPS and it relies on pre-extracted CLIP embeddings, optimizing only the geometric parameters of Gaussians rather than directly updating the embedded feature fields.
Achieving truly online SLAM with dense VLM feature optimization remains challenging.

First, extending Gaussian Splatting~\cite{3dgs} to handle high-dimensional feature vectors significantly increases computational complexity.
Furthermore, applying conventional alpha-blending to high-dimensional features is conceptually unsound. 
It blends semantics from multiple surfaces, producing ambiguous, non-interpretable feature vectors.
A similar issue has been implicitly recognized in object-centric methods such as ObjectGS \cite{objectgs}, which separate Gaussians by object to avoid semantic entanglement.

Additionally, the system faces memory inefficiency and inadequate compression. 
Storing high-dimensional features for millions of Gaussians is memory-intensive, so compression methods such as decoder is adopted~\cite{feature-3dgs,langsplat}.
This mitigates memory usage but degrades open-set segmentation performance and imposes a dual burden on the system, which must simultaneously reconstruct geometry and train a scene-specific decoder.
Moreover, this design tightly couples scene representation with a learned decoder, restricting generalization and downstream reasoning.

We present a language-aligned dense SLAM framework that resolves these issues by redesigning the entire rendering, map management, and optimization pipeline, achieving high geometric fidelity and semantic consistency while maintaining fast tracking and mapping. 
To meet strict time constraints, we adopt GS-ICP SLAM~\cite{gsicpslam} as our geometric backbone to utilize its fast pose tracking and Gaussian initialization scheme.
%
%
To address the computational and representational challenges, we adopt dual rendering strategies: conventional alpha-blending for the geometric field and Top-K rendering for the semantic field. Top-K selectively aggregates the most influential Gaussians per ray, improving efficiency and avoiding the semantic distortion inherent to alpha-blending, while alpha-blending preserves stable scene convergence. A custom CUDA kernel co-designed for Top-K further boosts this pipeline.

%
To mitigate the severe memory cost, we propose a multi-criteria map management that enforces semantic and geometric consistency.
A semantic-geometric consistency pruning only retains Gaussians that contributing to both color and feature renderings, ensuring representational consistency.
Second, geometric redundancy is eliminated by suppressing overlapping Gaussians during map updates, increasing compactness without additional computation.
%
Finally, we adopt a hybrid field optimization scheme to accelerate convergence under time constraints.
The feature field is smoother than the geometric field, and relies on a stable geometric structure.
We thus decouple optimization cycles to update geometry more frequently to ensure stability, and refine features at a lower rate on this geometric foundation.
This reduces redundant computation and speeds up convergence of both fields.

In summary, our contributions are as following:
\begin{itemize}
  \setlength{\itemsep}{0pt}
  \setlength{\topsep}{1pt}
    \item We introduce an online, language-aligned dense SLAM framework that constructs a 3D feature field directly from VLM embeddings, enabling open-vocabulary and LLM-interactive perception, while sustaining low-latency tracking and mapping.
    \item We adopt a dual rendering scheme, alpha-blending for geometric field and Top-K rendering for semantic field. This design drastically reduces computational overhead on a single GPU, enabling stable scene convergence, high-speed feature field updates while mitigating the semantic distortion inherent in alpha-blending.
    \item Taking into account the characteristics of dual rendering and the interdependence between geometric and semantic fields, we design a unified map management and hybrid optimization strategy that jointly ensures geometric–semantic consistency. The proposed pruning and dependency-aware scheduling reduce memory consumption and redundant computation, resulting in a compact, stable, and efficient mapping process.
\end{itemize}

\section{Related Work}
\label{sec:related_work}
\noindent\textbf{Semantic SLAM.}
As SLAM technology has advanced, efforts to integrate not only geometric structure but also semantic information into SLAM maps have evolved.
Object-oriented SLAM~\cite{salas2013slam,tian2021accurate,nicholson2018quadricslam,han2023sq,wang2024voom,zins2022oa,mccormac2018fusion++,yu2019variational,yang2019cubeslam,dube2020segmap} represents a scene in terms of objects and estimates each object’s attributes, and inter-object relationships, enabling higher-level representation.
While object-oriented SLAM enables high-level spatial understanding, it still faces label uncertainty and map sparsity. To address these issues, dense metric-semantic mapping has been explored~\cite{rosinol2020kimera,hughes2022hydra,yang2022sdf} 
With the advent of differentiable rendering~\cite{mildenhall2021nerf,3dgs}, SLAM systems can now reconstruct photo-realistic scenes~\cite{imap,coslam,eslam,nerfslam,gs-slam}. Building upon these advances, recent studies incorporated dense semantic information into scene. For instance, \cite{sni-slam} integrates RGB, depth, and semantic features into a NeRF-based framework through hierarchical semantic encoding. Furthermore, \cite{zhu2024semgauss,li2024sgs} introduced 3DGS-based semantic scene representations that embed semantic features within Gaussians and jointly optimize semantic-geometric fields.

\noindent\textbf{VLM Feature Embedded dense scene reconstruction.}
CLIP~\cite{clip} aligns visual and linguistic signals within a shared latent feature space, moving visual recognition beyond reliance on predefined, finite label sets toward an open-vocabulary paradigm. Building on this, LSeg~\cite{lseg} and OpenSeg~\cite{openseg} support pixel-wise inference of these features.
These advances naturally motivate extending open-vocabulary capability to 3D scene representations. One direction integrates language features into implicit neural fields~\cite{kerr2023lerf,li2024dns} by training networks that map 3D coordinates to density, color, and a language embedding at that location. More recently, LangSplat \cite{langsplat} and Feature-3DGS \cite{feature-3dgs} attach a language feature vector to each Gaussian in the scene, an explicit design that can accelerate rendering and facilitate editing.
\begin{figure*}[hbt!]
    \vspace{-10pt}
    \centering
    \includegraphics[width=\textwidth]{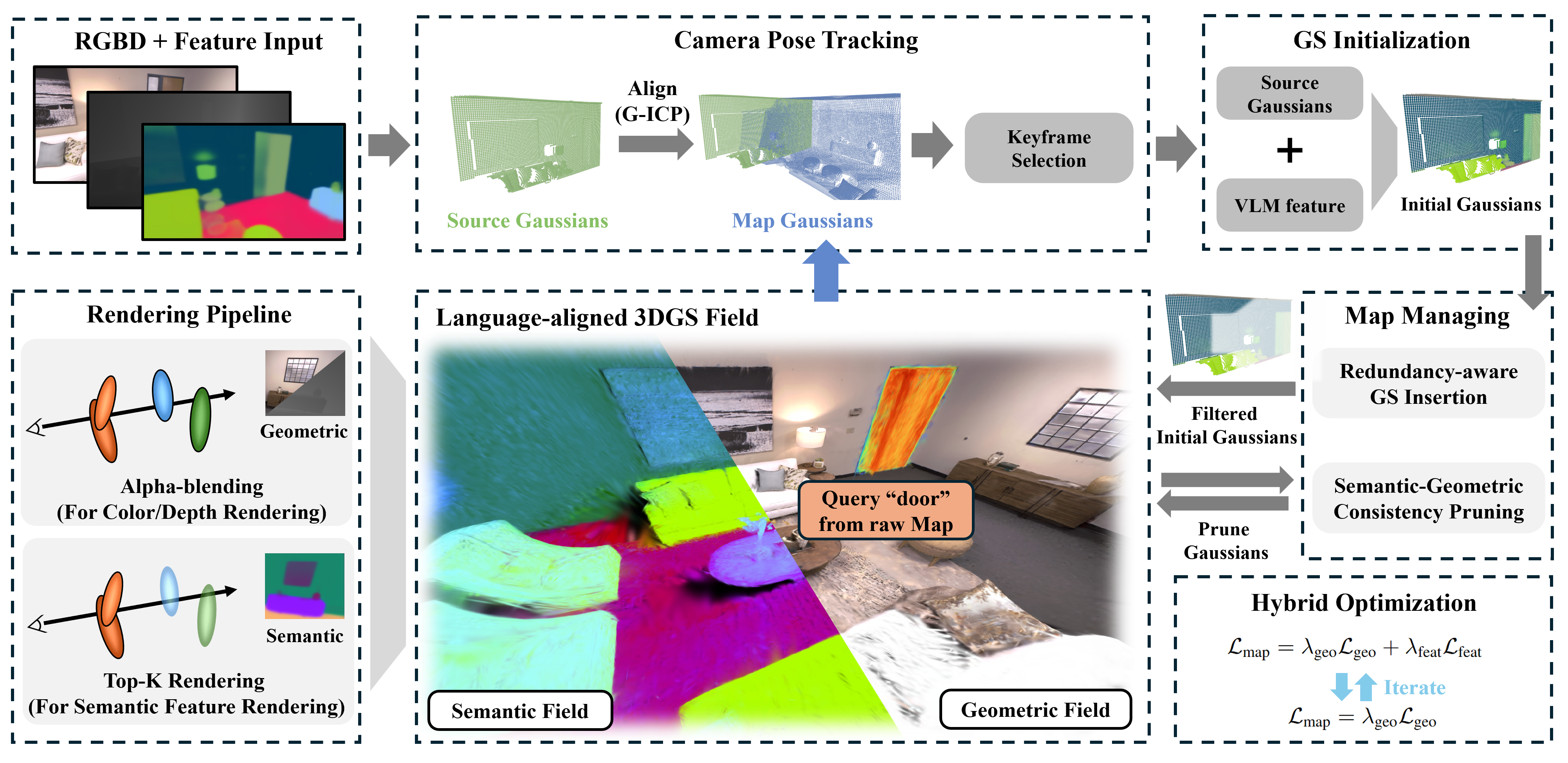}
    \caption{\textbf{Overview of the proposed SLAM framework.} 
        From RGB-D frames and VLM feature maps, the system constructs both geometric and semantic fields in real time.
        Source Gaussians are calculated from the depth input and aligned with existing map Gaussians via G-ICP to estimate camera poses.
        When a frame is selected as a keyframe, new Gaussians are initialized using geometric attributes obtained during tracking and feature vectors sampled from the input VLM feature map.
        A multi-criteria map management strategy prunes redundant Gaussians, reducing memory consumption and enforcing semantic–geometric consistency.
        Rendering is performed through two complementary schemes: alpha blending for geometry and Top-K rendering for semantics. 
        And the entire scene is jointly optimized using the proposed hybrid field optimization.}
    \label{fig:overview}
    \vspace{-20pt}
\end{figure*}

\section{Method}
Given RGB-D inputs and VLM features derived from LSeg~\cite{lseg}, we aim to construct a dense, language-aligned feature field in real time.
Our system operates through two parallel threads: tracking and mapping.

In the tracking thread, the camera pose is estimated by aligning source and map Gaussians using G-ICP~\cite{gicp}.
If the overlap ratio between them drops below a threshold, the current frame is selected as a keyframe, and its source Gaussians are initialized into the 3DGS scene.
Each new Gaussian is initialized with both geometric attributes and VLM features, facilitating fast map convergence.
Meanwhile, correspondence distances computed during G-ICP tracking are reused in the subsequent map management stage (Sec.~\ref{sec:map_managing}) to suppress redundant Gaussian insertion.

Concurrently, the mapping thread refines the global map through differentiable rendering and optimization.
We adopt Top-K rendering (Sec.~\ref{sec:topk}) to efficiently render high-dimensional features and a multi-criteria pruning strategy (Sec.~\ref{sec:map_managing}) to enforce geometric–semantic consistency with maintaining compactness.
Finally, a hybrid field optimization scheme (Sec.~\ref{sec:hybrid_field_opt}) jointly updates geometric and semantic fields under time constraints, achieving stable and efficient convergence.

\subsection{Scene Representation \& Geometric Rendering}
We model the scene as VLM feature embedded 3D Gaussian primitives. 
Each 3D Gaussian is parameterized with color, mean, covariance, opacity, and a VLM feature value. 
The rasterizer renders the geometric representation (color and depth) of the scene from these Gaussians. 
This process follows the alpha-blending method from \cite{3dgs}.
\begin{equation} \label{eqn:rendering}
    \mathbf{C}(\mathbf{p})
    =
    \sum_{i \in N} \mathbf{c}_i \alpha_i \prod^{i-1}_{j=1} (1-\alpha_j),\ 
    D(\mathbf{p})
    =
    \sum_{i \in N} z_i \alpha_i \prod^{i-1}_{j=1} (1-\alpha_j)
\end{equation}
%
$N$ is the number of Gaussians, $\mathbf{c}_i$ is the color of the Gaussian, $z_i$ is the z-depth of the Gaussian mean, and $\alpha_i$ is the Gaussian's opacity multiplied by the 2D Gaussian projected onto the view space. 
To achieve rapid scene convergence, we omit spherical harmonics (SH) to reduce complexity and achieve faster scene convergence. 
Thus, we retain the conventional alpha-blending method for color/depth rendering to ensure stable geometric reconstruction, while the Top-K rendering~\ref{sec:topk} is utilized for rendering VLM features.
\begin{figure}[t]
    \vspace{-10pt}
  \centering
   \includegraphics[width=0.75\linewidth]{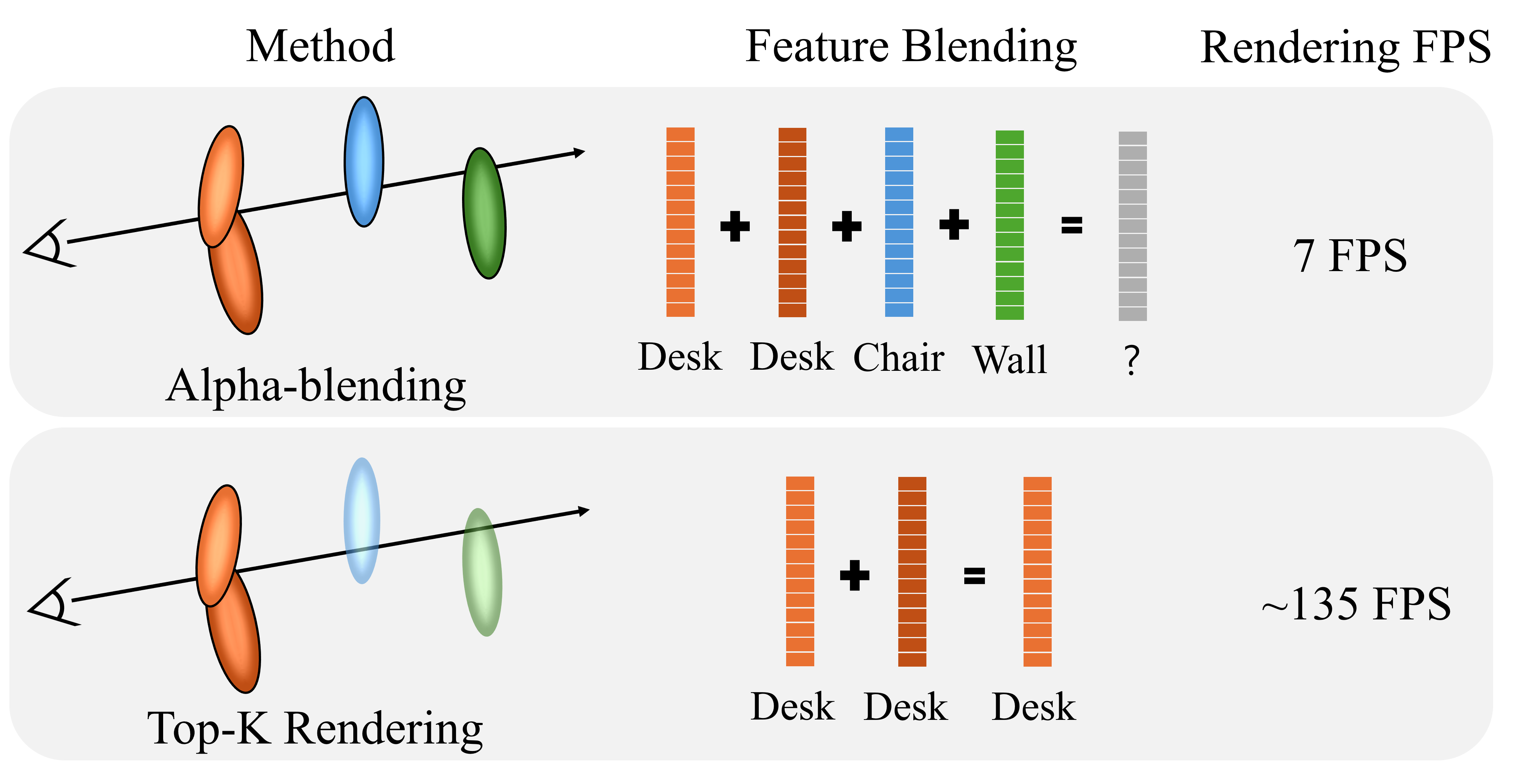}
   \caption{\textbf{Comparison between alpha-blending and the proposed Top-K rendering.} 
   Alpha-blending samples Gaussians off the surface, mixing unrelated features and incurring heavy cost by accumulating all ray-contributed high-dimensional features. In contrast, Top-K rendering aggregates only surface Gaussians, yielding consistent semantics and much higher efficiency.
   }
   \label{fig:topk}
   \vspace{-10pt}
\end{figure}
\subsection{Top-K Rendering for Semantic Feature} \label{sec:topk}
Rendering high-dimensional VLM features poses a significant computational challenge for SLAM systems.
While Feature3DGS~\cite{feature-3dgs} renders feature maps using the same alpha-blending scheme as color rendering, this approach introduces two key issues, heavy computational cost due to iteration over high-dimensional features, and the mixing of semantics from multiple surfaces.
To enable efficient and accurate feature rendering, we introduce Top-K method, a rendering mechanism tailored for online SLAM.

Our method identifies the $K$ most influential Gaussians for each ray based on their contribution weights computed during the alpha-blending process (Eqn.~\ref{eqn:rendering}).
Formally, the Top-K index set $\mathcal{K}$ is defined as:
\begin{equation} \label{eqn:topk_set} 
    \mathcal{K} = { \pi(1), ..., \pi(K) } \quad \text{s.t.} \quad w_{\pi(1)} \ge w_{\pi(2)} \ge \dots \ge w_{\pi(N)} 
\end{equation}
Since the semantic feature is fundamentally a unit vector representing a direction in the language space, the contributions of the selected Gaussians are re-normalized to compute the final weights $w'_k$.
\begin{equation} \label{eqn:renormalize} 
    w'_k = \frac{w_k}{\sum_{j \in \mathcal{K}} w_j} \quad \text{for} \quad k \in \mathcal{K} 
\end{equation}
%
Finally, the feature at pixel $\mathbf{p}$ is rendered as:
\begin{equation} \label{eqn:feature_rendering} 
    \mathbf{F}(\mathbf{p}) = \sum_{k \in \mathcal{K}} w'_k \mathbf{f}_k 
\end{equation}
where $\mathbf{f}_k$ denotes the VLM feature of Gaussian $\mathcal{G}_k$.
This formulation not only reduces rendering complexity but also naturally mitigates semantic distortion by focusing on the most dominant surface features.
To maximize the efficiency of this selective computation, we implement the entire process within a custom CUDA kernel designed for high-throughput rendering.
In our implementation, color/depth and semantic renderings are handled in separate kernels to enhance parallelism.
During color/depth rendering, the indices and blending weights of the Top-K Gaussians are recorded and reused in feature rendering kernel.
Since the number of contributing Gaussians per pixel is fixed to $K$, this design enables deterministic thread allocation and channel-wise parallel accumulation.
This enables high-speed rendering of both geometric and semantic fields. 

While we adopt Top-K for semantic features, we keep conventional alpha-blending for color to stabilize geometry. As noted in \cite{hahlbohm2025efficient}, applying Top-K to all fields can destabilize training and cause catastrophic forgetting.
Although this hybrid design provides the optimal rendering strategy for each field, it inevitably introduces minor inconsistencies between the geometric and semantic representations.
In particular, some Gaussians may contribute to color rendering but remain unused in feature rendering.
These inactive Gaussians tend to accumulate meaningless semantic features, that degrade map consistency.
In the following section, we describe how our proposed pruning strategy eliminates such artifacts and enforces a coherent geometric–semantic representation.
%

%
\subsection{Multi-Criteria Gaussian Managing for Compact and Consistent Representation} \label{sec:map_managing}
To tackle the heavy memory footprint of per-Gaussian VLM features and hybrid-rendering mismatches, we introduce a two-stage, multi-criteria pruning method that reduces redundancy while preserving geometric and semantic fidelity.
\vspace{-10pt}
\subsubsection{Semantic-Geometric Consistency Pruning}

The first mechanism refines the existing map to enforce semantic-geometric consistency and remove unnecessary Gaussians. 
Specifically, we measure the contribution of each Gaussian by counting the number of times it is selected for Top-K rendering. 
However, pruning based only on Top-K participation can remove Gaussians that are not selected at some viewpoints yet are crucial to the overall geometry, thereby harming geometric quality.
To solve this problem, we devised a 2-stage probabilistic pruning method that also considers geometric importance. First, we select Gaussians with low Top-K contributions as a primary candidate set ($\mathcal{G}_\text{prune}$). Then, within this candidate set, we calculate the maximum rendering contribution of each Gaussian as its geometric importance score $S_i$. This score is the maximum rendering contribution value computed for that Gaussian across all keyframes ($\mathcal{K}$) and rays ($\mathcal{R}$).
\begin{equation} \label{eqn:score} 
    S_i = \max_{k \in \mathcal{K}, \mathbf{r} \in \mathcal{R}_k} w_i(\mathbf{r})
\end{equation}
%
$w_i(\mathbf{r}) = \alpha_i(\mathbf{r}) \prod_{j=1}^{i-1} (1-\alpha_j(\mathbf{r}))$ denotes the rendering contribution calculated for alpha-blending (Eqn.\ref{eqn:rendering}).
Instead of simply removing Gaussians with low score, we normalize these scores to create a survival probability distribution $P_{\text{survive}}$.
\begin{equation} \label{eqn:pruning_prob} 
    P_{\text{survive}}(\mathcal{G}_i) = \frac{S_i}{\sum_{j \in \mathcal{\mathcal{G}}_{\text{prune}}} S_j} 
\end{equation}
%
Based on this probability distribution, we perform weighted sampling to keep a predefined ratio of Gaussians alive from the candidate set $\mathcal{G}_\text{prune}$.
This method achieves an effective balance, preserving the quality of both fields by giving geometrically important Gaussians a chance to survive, even if they are among those with low semantic contribution.
\vspace{-10pt}
\subsubsection{Redundancy-aware GS Insertion}
The second mechanism operates preemptively during the new Gaussian insertion stage, preventing unnecessary redundancy in the map from the start. 
It determines if the area where a new Gaussian is to be added is already sufficiently represented by the existing map, and suppresses the addition if redundancy is anticipated.

To maximize the efficiency of this process, we reuse the correspondence distance information already computed during the G-ICP tracking stage. 
If the distance between a source Gaussian and its nearest target Gaussian is below a certain threshold, we consider the area already well-represented and do not add the new Gaussian. 
This method effectively controls map redundancy with no additional computational cost.

\subsection{Hybrid Field Optimization} \label{sec:hybrid_field_opt}

With our efficient rendering method (Sec.~\ref{sec:topk}) and map management strategy (Sec.~\ref{sec:map_managing}), the final objective is to ensure rapid and stable convergence of the high-dimensional, multi-field representation under time constraints.
All parameters of the map Gaussians (geometric attributes $\mathcal{G}$ and semantic attributes $\mathbf{f}$) are optimized in the mapping thread by minimizing the difference between rendered outputs and ground-truth keyframe images.
The overall loss is defined as a weighted sum of geometric and semantic reconstruction losses:
\begin{equation} \label{eqn:total_loss}
    \mathcal{L}_{\text{map}} = \lambda_{\text{geo}} \mathcal{L}_{\text{geo}} + \lambda_{\text{feat}} \mathcal{L}_{\text{feat}}
\end{equation}
The geometric reconstruction loss supervises both color and depth consistency:
\begin{equation} \label{eqn:geo_loss}
    \mathcal{L}_{\text{geo}} = (1 - \lambda_1)\mathcal{L}_1(\mathbf{C}, \mathbf{C}_{gt}) + \lambda_1\mathcal{L}_1(\mathbf{C}, \mathbf{C}_{gt}) + \lambda_{\text{2}}\mathcal{L}_1(D, D_{gt})
\end{equation}
%
where $\mathbf{C}_\text{gt}$ and $D_\text{gt}$ denote ground-truth color and depth images.
The semantic reconstruction loss optimizes the Gaussian feature $\mathbf{f}$ using the L1 distance between the rendered feature map $\mathbf{F}$—obtained through Top-K rendering (Eqn.~\ref{eqn:feature_rendering})—and the ground-truth VLM feature map $\mathbf{F}_{gt}$:
\begin{equation} \label{eqn:feat_loss}
    \mathcal{L}_{\text{feat}} = \mathcal{L}_1(\mathbf{F}, \mathbf{F}_\text{gt})
\end{equation}
%

Optimizing geometric and semantic fields at every iteration is costly, even with an efficient renderer. The geometric field carries high-frequency structure, while the semantic field varies more smoothly and depends on stable geometry; updating semantics on moving geometry yields inefficient, suboptimal learning.

We therefore adopt Hybrid Field Optimization: an asymmetric schedule that decouples update rates. Geometry—the structural backbone—is updated more frequently to stabilize the map; semantics are then refined at a lower rate on this foundation. This cuts redundant computation, speeds convergence, and improves overall system stability.

\section{Experiments}
In this section, we validate the performance of our proposed method and present ablation results to demonstrate the effectiveness of our proposed modules.
%
%
\subsection{Experimental Setup}
\textbf{Datasets.}
We evaluate on Replica~\cite{replica_dataset} and TUM-RGBD~\cite{tum_dataset}. Replica provides high-quality synthetic RGB-D data, whereas TUM-RGBD offers real-world sequences with substantial noise and frequent depth missing regions. By utilizing both datasets with these different characteristics, we demonstrate that our method operates robustly in both ideal and challenging, noisy real-world environments.

\noindent\textbf{Implementation Details.}
All experiments were performed on a desktop with a Ryzen9 7900x CPU and an NVIDIA RTX 4090 GPU (24GB RAM).
Our target is achieving online semantic SLAM, hence we avoid complex pre-processing used by LeRF~\cite{kerr2023lerf} and LangSplat~\cite{langsplat} for supervision feature maps.
Instead, following Feature3DGS~\cite{feature-3dgs}, we obtain the ground-truth (GT) feature maps directly from LSeg and use them as supervision.
We applied Semantic–Geometric Consistency Pruning every 500 iterations, resampling 50\% Gaussians of $\mathcal{G}_\textbf{prune}$ to remove.
To enhance tracking accuracy, we refine camera poses using rendering-loss based pose refinement~\cite{monogs} during map optimization.

\noindent\textbf{Metrics.}
Tracking accuracy is measured by Absolute Trajectory Error (ATE) RMSE, and geometric fidelity is evaluated with PSNR, SSIM, and LPIPS.
Semantic fidelity is measured using pixel-wise accuracy and mIoU (mean Intersection over Union), following Feature3DGS.
For all offline baselines~\cite{kerr2023lerf,langsplat,feature-3dgs} and our method, we use the same LSeg-derived GT feature maps to ensure a consistent evaluation protocol.
For system fps, FPS is computed from the total elapsed time of the full pipeline for SLAM methods. 
For offline methods, it is computed from the total reconstruction time excluding COLMAP-based pose estimation~\cite{sfm}.
We evaluated rendered images of keyframes and the reported results are the best values of 3 runs.

\noindent\textbf{Baselines.}
To compare the performance of our system from multiple perspectives, we selected two groups of baselines.
Tracking and Geometric Fidelity were evaluated against existing state-of-the-art NeRF and 3DGS-based SLAM methods \cite{nice-slam,point-slam,monogs,splatam}.
For Semantic Fidelity, there is a scarcity of comparable open-set SLAM approaches. Therefore, we performed a comparative evaluation against recent offline reconstruction techniques, LeRF \cite{kerr2023lerf}, LangSplat \cite{langsplat} and Feature3DGS \cite{feature-3dgs}. 
\begin{table*}[hbt!]
\vspace{-20pt}
  \caption{\textbf{Tracking Accuracy and Geometric fidelity} 
  Our method attains the best tracking accuracy on Replica and competitive accuracy on TUM-RGBD, while maintaining high system (tracking and mapping) speed. Despite jointly optimizing both geometric and semantic fields, our system surpasses geometry-only baselines in both speed and geometric quality. All results are reported without any post-optimization.
  }
  \centering
  \resizebox{\textwidth}{!}{%
    \begin{tabular}{cccccc|ccccc}
    \toprule
    \multirow{2}{*}{Method} & \multicolumn{5}{c}{Replica Dataset} & \multicolumn{5}{c}{TUM-RGBD Dataset} \\
    \cmidrule(lr){2-11}
    & PSNR [dB] $\uparrow$  & SSIM $\uparrow$  & LPIPS $\downarrow$ & ATE RMSE [cm] $\downarrow$ & System FPS $\uparrow$ & PSNR [dB] $\uparrow$  & SSIM $\uparrow$  & LPIPS $\downarrow$ & ATE RMSE [cm] $\downarrow$ & System FPS $\uparrow$ \\
    \midrule
    Point-SLAM~\cite{point-slam} &\secondbg{35.56}&\bestbg{0.977}&\thirdbg{0.118}&0.471&0.415&\thirdbg{21.33}&\thirdbg{0.733}&0.453&\thirdbg{2.517}&0.254\\
    SplaTAM~\cite{splatam} &\thirdbg{34.19}&\secondbg{0.970}&\bestbg{0.087}&\thirdbg{0.367}&\secondbg{1.184}&\secondbg{23.53}&\bestbg{0.909}&\secondbg{0.166}&3.263& \thirdbg{1.184}\\
    MonoGS~\cite{monogs} &35.34&0.944&0.122&\secondbg{0.318}&\thirdbg{0.679}&18.07&0.726&\thirdbg{0.320}&\bestbg{1.520}&\secondbg{2.283} \\
    \textbf{Ours} &\bestbg{35.92}&\thirdbg{0.952}&\secondbg{0.099}&\bestbg{0.213}&\bestbg{15}&\bestbg{23.78}&\secondbg{0.856}&\bestbg{0.147}&\secondbg{2.316}&\bestbg{15} \\
    \bottomrule
    \end{tabular}%
  }
  \label{tab:slam_eval}
  \vspace{-20pt}
\end{table*}
\subsection{Tracking Accuracy}
Tab.\ref{tab:slam_eval} reports the tracking accuracy on the Replica and TUM-RGBD datasets.
Our method achieves state-of-the-art tracking on Replica and competitive results on TUM-RGBD. 
Replica’s high-fidelity RGB-D places an upper bound on achievable accuracy, and our results indicate that the proposed pipeline fully exploits such inputs. 
On the noisier TUM-RGBD, despite our baseline’s reliance on depth-based G-ICP, rendering-loss–based pose refinement improves robustness by leveraging photometric consistency.
\vspace{-20pt}
\subsection{Quality of the Reconstructed Scene}
Since our system aims to reconstruct both geometric and semantic fields, we evaluate the reconstructed scenes from these two complementary perspectives.

\noindent\textbf{Geometric Fidelity.}
To evaluate geometric reconstruction fidelity, we used the standard rendering metrics, comparing our method against recent NeRF-based \cite{point-slam}, and 3DGS-based \cite{monogs,splatam} SLAM baselines.
Note that our system performs a fundamentally more complex task, as it must simultaneously reconstruct both the geometric field and a high-dimensional semantic field, whereas the baselines focus solely on geometric reconstruction. 
Despite this inherent complexity, our method achieves outstanding map quality on both Replica and TUM datasets while maintaining fast tracking and mapping speed as shown in Tab.~\ref{tab:slam_eval}.
Remarkably, our system achieves significantly higher tracking and mapping FPS than competing methods without sacrificing geometric accuracy.
This stems from the optimized rendering pipeline (\ref{sec:topk}), which eliminates the bottleneck of high-dimensional optimization, and the Hybrid Field Optimization strategy (\ref{sec:hybrid_field_opt}), which prioritizes rapid geometric convergence.
As a result, our method demonstrates that online high-dimensional semantic mapping can coexist with geometric precision.

\begin{table}[hbt!]
    \vspace{-15pt}
  \caption{\textbf{Evaluation Results of Semantic Fidelity on Replica Dataset.} The proposed method demonstrates higher semantic fidelity than LeRF and LangSplat, and comparable performance to Feature3DGS while delivering fast system speed.}
  \centering
  \resizebox{0.8\columnwidth}{!}{%
    \begin{tabular}{cccccccccccc}
    \toprule
    Method         & Metric     & r0 & r1 & r2 & o0 & o1 & o2 & o3 & o4 & Avg. \\
    \midrule
        \multirow{3}{*}{LeRF \cite{kerr2023lerf}} 
            & Accuracy $\uparrow$ & 0.494 & 0.697& 0.710& 0.633& 0.613& 0.557& 0.554& 0.685& 0.618 \\
            & mIoU $\uparrow$ & 0.272& 0.217& 0.358& 0.362& 0.323& 0.150& 0.201& 0.333& 0.277 \\
            & FPS $\uparrow$ & \secondbg{5.376} & \secondbg{5.323} & \secondbg{5.368} & \secondbg{5.402} & \secondbg{5.427} & \secondbg{5.413} & \secondbg{5.428} & \secondbg{5.403} & \secondbg{5.392} \\
    \midrule
        \multirow{3}{*}{LangSplat \cite{langsplat}} 
            & Accuracy $\uparrow$   & 0.544 & 0.549& 0.701& 0.345& 0.655& 0.772& 0.651& 0.690& 0.614 \\
            & mIoU $\uparrow$       & 0.264& 0.184& 0.330& 0.125& 0.227& 0.375& 0.289& 0.311& 0.263 \\
            & FPS $\uparrow$        & 1.047 & 0.794 & 1.049 & 0.570 & 0.571 & 0.693 & 1.080 & 1.101 & 0.863 \\
    \midrule
        \multirow{3}{*}{\shortstack{Feature 3DGS \cite{feature-3dgs}\\(512\text{-}D)}}
            & Accuracy $\uparrow$ & \secondbg{0.926} & \bestbg{0.951}& \bestbg{0.922}& \bestbg{0.744}& \bestbg{0.796}& \bestbg{0.954}& \bestbg{0.940}& \bestbg{0.910}& \bestbg{0.893} \\
            & mIoU $\uparrow$ & \secondbg{0.836}& \bestbg{0.718}& \secondbg{0.738}& \secondbg{0.546}& \secondbg{0.434}& \bestbg{0.687}& \secondbg{0.775}& \secondbg{0.633}& \secondbg{0.671} \\
            & FPS $\uparrow$ & 0.242 & 0.245 & 0.276 & 0.337 & 0.373 & 0.296 & 0.292 & 0.335 & 0.300 \\
    \midrule
        \multirow{3}{*}{\shortstack{Feature 3DGS \cite{feature-3dgs}\\(128\text{-}D)}}
            & Accuracy $\uparrow$ & \bestbg{0.928} & \secondbg{0.949}& \secondbg{0.917}& \secondbg{0.743}& \thirdbg{0.785}& \secondbg{0.953}& \secondbg{0.939}& 0.907& \secondbg{0.890} \\
            & mIoU $\uparrow$ & \bestbg{0.838}& \secondbg{0.712}& \thirdbg{0.729}& \thirdbg{0.538}& \thirdbg{0.406}& \thirdbg{0.677}& \thirdbg{0.763}& \thirdbg{0.615}& \thirdbg{0.660} \\
            & FPS $\uparrow$ & \thirdbg{1.143} & \thirdbg{1.182} & \thirdbg{1.360} & \thirdbg{1.591} & \thirdbg{1.723} & \thirdbg{1.421} & \thirdbg{1.455} & \thirdbg{1.584} & \thirdbg{1.432} \\
    \midrule
        \multirow{3}{*}{\textbf{Ours}}
            & Accuracy $\uparrow$ & \thirdbg{0.904} & \thirdbg{0.939} & \thirdbg{0.916} & \bestbg{0.744} & \secondbg{0.793} & \thirdbg{0.934} & \thirdbg{0.925} & \secondbg{0.908} & \thirdbg{0.883} \\
            & mIoU $\uparrow$   & \thirdbg{0.800} & \thirdbg{0.711} & \bestbg{0.747} & \bestbg{0.549} & \bestbg{0.440} & \secondbg{0.683} & \bestbg{0.776} & \bestbg{0.677} & \bestbg{0.673} \\
            & FPS $\uparrow$ & \bestbg{15} & \bestbg{15} & \bestbg{15} & \bestbg{15} & \bestbg{15} & \bestbg{15} & \bestbg{15} & \bestbg{15} & \bestbg{15} \\
    \bottomrule
    \end{tabular}%
  }
  \label{tab:semantic_fidelity_replica}
  \vspace{-15pt}
\end{table}

\begin{table}[hbt!]
\vspace{-15pt}
  \caption{\textbf{Evaluation Results of Semantic Fidelity on TUM-RGBD Dataset.} Our method consistently maintains semantic fidelity, demonstrating robustness against sensor noise.}
  \centering
  \resizebox{0.6\columnwidth}{!}{%
    \begin{tabular}{cccccc}
    \toprule
    Method         & Metric     & fr1/desk & fr2/xyz & fr3/desk & Avg. \\
    \midrule
        \multirow{3}{*}{LeRF \cite{kerr2023lerf}} 
            & Accuracy $\uparrow$ & 0.716 & 0.130 & 0.624 & 0.490\\
            & mIoU $\uparrow$ & 0.310 & 0.036 & 0.444 & 0.263 \\
            & FPS $\uparrow$ & \secondbg{2.423} & \secondbg{10.250} & \secondbg{7.493} & \secondbg{6.722} \\
    \midrule
        \multirow{3}{*}{LangSplat \cite{langsplat}} 
            & Accuracy $\uparrow$ & 0.640 & 0.499 & 0.493 & 0.544 \\
            & mIoU $\uparrow$ & 0.250 & 0.199 & 0.237 & 0.229 \\
            & FPS $\uparrow$ & 0.512 & 0.322 & 0.202 & 0.345 \\
    \midrule
        \multirow{3}{*}{\shortstack{Feature 3DGS \cite{feature-3dgs}\\(512\text{-}D)}}
            & Accuracy $\uparrow$ & \bestbg{0.868} & \bestbg{0.849} & \secondbg{0.789} & \bestbg{0.835} \\
            & mIoU $\uparrow$ & \bestbg{0.615} & \secondbg{0.641} & \secondbg{0.644} & \secondbg{0.633} \\
            & FPS $\uparrow$ & 0.163 & 0.790 & 0.614 & 0.522 \\
    \midrule
        \multirow{3}{*}{\shortstack{Feature 3DGS \cite{feature-3dgs}\\(128\text{-}D)}}
            & Accuracy $\uparrow$ & \secondbg{0.865} & \secondbg{0.847} & \thirdbg{0.788} & \secondbg{0.833} \\
            & mIoU $\uparrow$ & \thirdbg{0.594} & \thirdbg{0.639} & \thirdbg{0.641} & \thirdbg{0.625} \\
            & FPS $\uparrow$ & \secondbg{0.815} & \secondbg{4.406} & \secondbg{3.262} & \secondbg{2.828} \\
    \midrule
        \multirow{3}{*}{\textbf{Ours}}
            & Accuracy $\uparrow$ & \thirdbg{0.838} & \thirdbg{0.842} & \bestbg{0.805} & \thirdbg{0.828} \\
            & mIoU $\uparrow$   & \secondbg{0.600} & \bestbg{0.658} & \bestbg{0.667} & \bestbg{0.642} \\
            & FPS $\uparrow$ & \bestbg{15} & \bestbg{15} & \bestbg{15} & \bestbg{15} \\
    \bottomrule
    \end{tabular}%
  }
  \label{tab:semantic_fidelity_tum}
  \vspace{-15pt}
\end{table}

\noindent\textbf{Semantic Fidelity.}
To evaluate the semantic fidelity of our reconstruction scene, we compare against state-of-the-art offline open-set methods~\cite{kerr2023lerf,langsplat,feature-3dgs}.
Because these baselines operate offline with precomputed camera poses~\cite{sfm} and no runtime constraints, they represent a practical quality upper bound under our evaluation protocol.
We additionally report the system FPS of these methods based on total elapsed time.

LeRF’s NeRF-based volumetric field often smooths boundaries, which can limit dense, pixel-aligned semantics.
LangSplat compresses features into a very low-dimensional latent using a pretrained autoencoder. 
While this is memory-efficient, it can reduce fine-grained, pixel-level cues.
Consistently, Feature3DGS-128D underperforms Feature3DGS-512D, indicating that compression tends to reduce VLM embedding richness by discarding high-frequency semantics.

Our method outperforms LeRF and LangSplat in pixel accuracy and mIoU, and attains comparable pixel accuracy while higher mIoU than Feature3DGS.
This suggests that while minor deviations may appear in background regions, our reconstructions yield sharper object boundaries and more stable segmentations.
We attribute this to the Top-K rendering, which selectively aggregates surface-aligned semantic features and mitigates feature blending artifacts.
\subsection{Ablation Study}
\begin{table}[hbt!]
\vspace{-15pt}
  \caption{\textbf{Ablation Results on Top-K Rendering. } Our optimized rasterization pipeline improves rendering speed and SLAM optimization efficiency, with $K=3$ achieving the best trade-off between speed and semantic stability.}
  \label{tab:topk_rendering_ablation}
  \centering
  \resizebox{0.6\columnwidth}{!}{
    \begin{tabular}{c|cccccc}
      \toprule
      Method & PSNR & SSIM & LPIPS & Accuracy & IoU & Rendering FPS \\ 
      \midrule
      vanilla & 23.23 & 0.711 & 0.439 & 0.874 & 0.653 & 7 \\
      top-10 & 32.24 & 0.925 & 0.151 & 0.883 & 0.669 & 90 \\
      top-5 & 34.15 & 0.938 & 0.127 & 0.885 & \textbf{0.673} & 103 \\
      top-3 & 34.52 & 0.941 & 0.121 & \textbf{0.886} &\textbf{0.673} & 122 \\
      top-1 & \textbf{34.94} & \textbf{0.946} & \textbf{0.111} & 0.882 & 0.667 & \textbf{135} \\
      \bottomrule
    \end{tabular}
  }
  \vspace{-15pt}
\end{table}
\noindent\textbf{Ablation on Top-K Rendering.}
Tab.~\ref{tab:topk_rendering_ablation} shows the ablation results on the Top-K rendering.
The vanilla (full alpha-blending) baseline fails in both geometry and semantics due to a severe rendering bottleneck that limits optimization steps.
In contrast, Top-K rendering increases both geometric and semantic fidelity, while the value of K reveals a trade-off relationship. Increasing K decreases the rendering speed, degrades geometric quality. So K=1 shows the best geometric fidelity, but semantic fidelity of this setting is lower than that of K=3 or K=5.
We attribute this to the fact that when K is extremely small, rendering becomes sensitive to errors. If the single highest-contribution Gaussian is an artifact or the G.T. feature contains noise, this sensitivity leads to instability. In contrast, K=3 or K=5 averages out such noise by blending a small local neighborhood of high-contributing Gaussians. This robustness leads to a more stable semantic rendering. Conversely, we also observed that setting K too large (K=10) causes the Top-K rendering to approximate alpha-blending, reintroducing the semantic ambiguity thus decreasing semantic fidelity.
Considering the trade-off between rendering speed and stable semantic fidelity, we adopt the Top-3 as our final design.

\begin{table}[hbt!]
    \vspace{-15pt}
  \caption{\textbf{Ablation Results on Convergence of Field.} Reported results are average of replica 8 scenes and 3 scenes of TUM-RGBD dataset.}
  \label{tab:system_speed_ablation}
  \centering
  \resizebox{0.6\columnwidth}{!}{
    \begin{tabular}{c|cc|ccccc}
    \toprule
        Dataset & Top-K & Hybrid & PSNR & SSIM & LPIPS & Accuracy & mIoU \\ 
    \midrule
        \multirow{3}{*}{Replica}
        &\xmark & \xmark & 23.23 & 0.711 & 0.439 & 0.874 & 0.653 \\
        &\checkmark & \xmark & 34.48 & 0.941 & 0.121 & \textbf{0.883} & 0.672 \\
        &\checkmark & \checkmark & \textbf{35.92} & \textbf{0.952} & \textbf{0.099} & \textbf{0.883} & \textbf{0.673} \\
    \midrule
        \multirow{3}{*}{TUM-RGBD}
        &\xmark & \xmark & 17.60 & 0.687 & 0.332 & 0.782 & 0.578 \\
        &\checkmark & \xmark & 22.13 & 0.825 & 0.178 & 0.823 & 0.631 \\
        &\checkmark & \checkmark & \textbf{23.78} & \textbf{0.856} & \textbf{0.147} & \textbf{0.828} & \textbf{0.642} \\
    \bottomrule
    \end{tabular}
  }
  \vspace{-15pt}
\end{table}

\noindent\textbf{Ablation on Convergence of Field.}
%
%
%
We performed an ablation study to validate how our proposed modules accelerate field convergence. 
As shown in Tab. \ref{tab:system_speed_ablation}, our optimized rendering pipeline, substantially accelerates scene convergence. This is clearly confirmed by the PSNR improvement of approximately 1.5x on the Replica dataset and 1.3x on the TUM-RGBD dataset.
%
And hybrid optimization further improved PSNR on both datasets, demonstrating it is an effective strategy for enhancing geometric fidelity of the scene.
The more compelling results emerged from the semantic fidelity analysis.
On the Replica dataset containing high-quality images, the use of hybrid optimization showed negligible difference in semantic performance. 
We attribute this to the high-quality sensor data and our efficient Top-K renderer, which allowed the geometric field to converge rapidly even without the hybrid strategy.
Conversely, on the noisy TUM-RGBD dataset, applying hybrid optimization yielded a significant improvement in semantic fidelity.

This contrast provides experimental evidence for our hypothesis. In realistic, noisy environments where the geometric field struggles to converge stably, our hybrid optimization first forces this unstable geometric field to stabilize. This clearly demonstrates that the dependent semantic field can only be optimized correctly after this stable geometric foundation is established.
\begin{table}[hbt!]
\vspace{-15pt}
  \caption{\textbf{Ablation Results on Map Managing.} 
  Without map management, the system fails on both datasets due to memory exhaustion. Our methods reduce the Gaussian count by approximately 30\% while maintaining comparable rendering quality.}
  \label{tab:pruning_ablation}
  \centering
  \resizebox{0.9\columnwidth}{!}{
    \begin{tabular}{c|cc|ccccccc}
      \toprule
        Dataset & Redundancy & Top-K & PSNR & SSIM & LPIPS & Accuracy & mIoU & Num\_GS & Memory Usage\\ 
      \midrule
          \multirow{3}{*}{Replica}
          &\xmark & \xmark & x & x & x & x & x & 912595 & $>$ 24GB\\
          &\checkmark & \xmark & \textbf{36.30} & \textbf{0.955} & \textbf{0.091} & \textbf{0.882} & \textbf{0.667} & 281927 & 8.5GB\\
          &\checkmark & \checkmark & 35.94 & 0.953 & 0.098 & \textbf{0.882} & \textbf{0.667} & \textbf{215857} & \textbf{7.7GB}\\
      \midrule
          \multirow{3}{*}{TUM-RGBD}
          &\xmark & \xmark & x & x & x & x & x & 971818 & $>$ 24GB\\
          &\checkmark & \xmark & 23.21 & 0.844 & 0.155 & \textbf{0.828} & 0.641 & 211317 & 5.6GB\\
          &\checkmark & \checkmark & \textbf{23.78} & \textbf{0.856} & \textbf{0.147} & \textbf{0.828} & \textbf{0.642} & \textbf{91202} & \textbf{4.4GB}\\
      \midrule
          \multirow{3}{*}{TUM-RGBD, fr-1}
          &\xmark & \xmark & 20.52 & 0.801 & 0.218 & 0.833 & 0.598 & 371014 & 6.9GB\\
          &\checkmark & \xmark & 20.86 & 0.807 & 0.212 & \textbf{0.838} & \textbf{0.600} & 198827 & 5.3GB\\
          &\checkmark & \checkmark & \textbf{22.06} & \textbf{0.834} & \textbf{0.183} & \textbf{0.838} & \textbf{0.600} & \textbf{117295} & \textbf{4.6GB}\\
      \bottomrule
    \end{tabular}
  }
  \vspace{-10pt}
\end{table}


\noindent\textbf{Ablation on Map Managing.}
Tab.~\ref{tab:pruning_ablation} analyzes the impact of our map management modules.
Without pruning and redundancy control, the system runs out of GPU memory due to the uncontrolled growth of Gaussians, especially on long sequences such as Replica (2000 frames) and TUM-RGBD fr-2 (3397 frames) and fr-3 (2515 frames).
We report only the number of Gaussians by conducting additional experiments without VLM feature embeddings for this case.
Without them, the system ran out of memory due to the excessive number of Gaussians on most scenes.
Only the TUM fr-1 sequence (about 600 frames) was operable without map management, where we report the full metrics.

On the Replica dataset, pruning slightly decreases PSNR while reducing the number of Gaussians by 76.3\%.
Our pruning mechanism is intentionally designed to balance compression and fidelity by removing redundant or visually insignificant Gaussians.
Because high-quality synthetic scenes encourage the model to capture extremely fine details, a minor PSNR degradation occurs as high-frequency components are selectively pruned for compactness.

In contrast, on the real-world dataset~\cite{tum_dataset}, pruning not only maintains but even improves visual metrics while the Gaussian count drops by 90.6\%.
This clearly demonstrates the artifact-removal ability of our pruning mechanism.
Noisy depth inputs often lead to “floating” or unstable Gaussians unrelated to the true scene geometry, and our pruning eliminates these artifacts while preserving the essential geometry of the map.
Consequently, our method achieves higher scene fidelity with lower memory usage, confirming that map management effectively control the number of Gaussians while maintaining quality.
\begin{table}[hbt!]
    \vspace{-15pt}
  \caption{\textbf{Evaluation of Semantic-Geometric Consistency Pruning in Novel Views.} Our method improves both semantic and geometric fidelity in novel view rendering by removing ambiguous and inconsistent Gaussians.}
  \label{tab:novel_view}
  \centering
  \resizebox{0.55\columnwidth}{!}{
    \begin{tabular}{c|c|ccccc}
      \toprule
      Scene & Pruning & PSNR & SSIM & LPIPS & Accuracy & mIoU \\ 
      \midrule
      \multirow{2}{*}{room2}   & Without & 17.05 & 0.798 & 0.241 & 0.749 & 0.393 \\
                            &With & \textbf{17.16} & \textbf{0.802} & \textbf{0.237} & \textbf{0.753} & \textbf{0.397} \\
      \midrule
      \multirow{2}{*}{office4}   & Without & 16.84 & 0.821 & 0.239 & 0.735 & 0.367 \\
                                 &With &  \textbf{16.88} &  \textbf{0.824} &  \textbf{0.234} &  \textbf{0.740} &  \textbf{0.368} \\
      \bottomrule
    \end{tabular}
  }
  \vspace{-10pt}
\end{table}

\noindent\textbf{Effect of Semantic-Geometric Consistency Pruning on Novel Views.}
%
%
Tab.~\ref{tab:novel_view} shows the effect of our Semantic-Geometric Consistency Pruning on novel views. We used images from 200 random viewpoints in the Replica dataset, and held Redundancy-aware GS insertion constant and compared the results solely based on the application of Semantic-Geometric Consistency Pruning.

With pruning, all rendering metrics consistently improves compared to the baseline without it.
This demonstrates that our proposed pruning mechanism successfully removes the semantic artifacts, just as intended. 
As a result, the removal of these semantically ambiguous Gaussians led to an improvement in novel view semantic fidelity.
Furthermore, the experiment reveals a significant additional benefit: geometric fidelity improves concurrently. 
This suggests that the removed artifacts were not only semantically ambiguous but were also acting as geometric floaters that degraded the quality of novel view rendering.
In conclusion, our Semantic-Geometric Consistency Pruning goes beyond simple map compression. It functions as a crucial regularization mechanism that resolves the discrepancy-induced artifacts from our hybrid rendering approach, thereby improving both the geometric and semantic integrity of the final map.
\begin{figure}[hbt!]
\vspace{-15pt}
  \centering
   \includegraphics[width=0.9\linewidth]{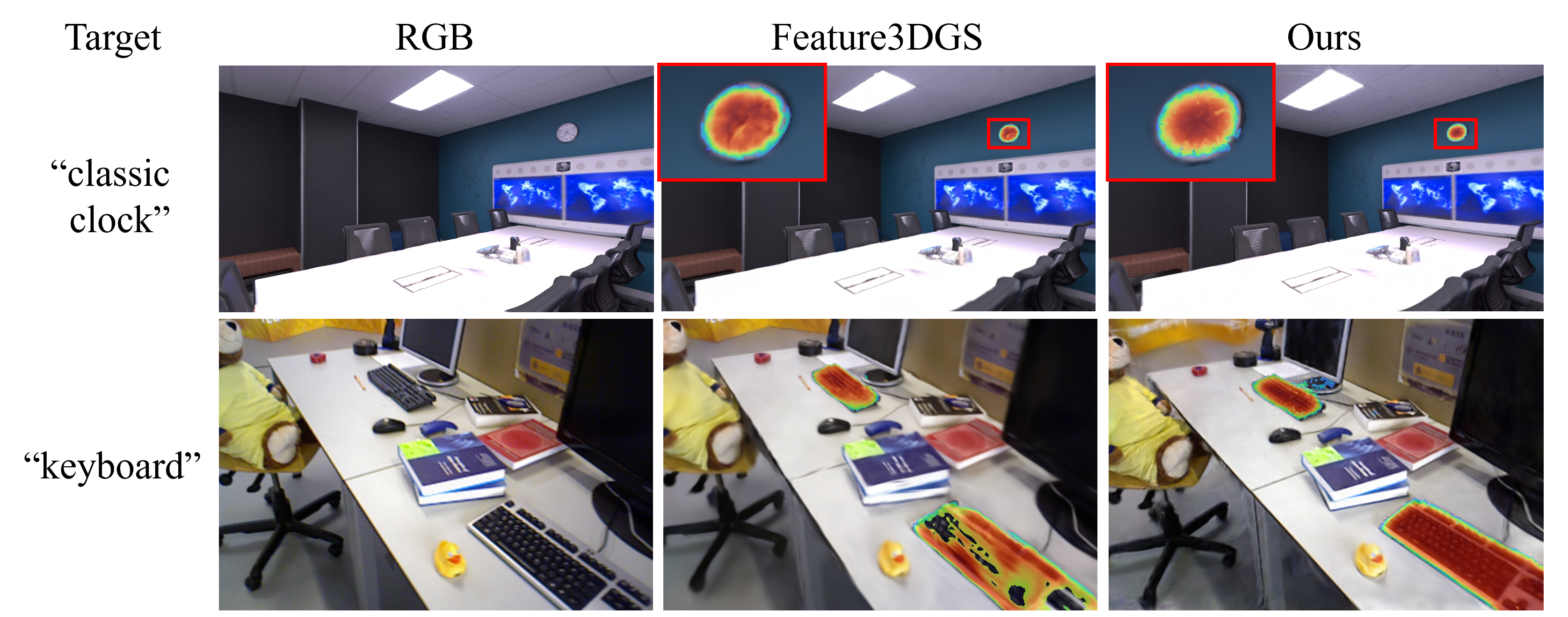}
   \caption{\textbf{Qualitative comparison with Offline Method.}
   The proposed method delivers text-query segmentation results comparable to offline approach.
   Furthermore, Top-K rendering and our pruning suppress noisy Gaussians, yielding robust segmentation.
   }
   \vspace{-15pt}
   \label{fig:vis_target}
\end{figure}

\noindent\textbf{Qualitative comparison with offline method.}
Fig.~\ref{fig:vis_target} shows a qualitative comparison of text-query segmentation against the offline baseline~\cite{feature-3dgs}. 
For a fair comparison, we evaluate Feature3DGS in its uncompressed configuration, which serves as its best performance.
Our method delivers segmentation quality comparable to the offline baseline.
Interestingly, our method shows more robust segmentation results, especially in real-world datasets with enormous sensor noise.
we hypothesize that Top-K rendering inherently suppresses the influence of noisy Gaussians during rendering, while our semantic–geometric consistency pruning explicitly removes them.

\section{Conclusion}
We presented a real-time, language-aligned dense feature field SLAM system.
Top-K rendering pipeline designed for real-time SLAM system removes the computational bottleneck of high-dimensional feature rendering and mitigates the semantic ambiguity inherent in alpha-blending.
Our map-management modules reduces semantic-geometric discrepancy and minimize redundant Gaussians.
With our extensive experiments, proposed modules showed their effectiveness.
Overall, our system runs at 15FPS, surpasses geometric-only state-of-the-art methods in geometric fidelity, and achieves comparable semantic fidelity to offline approaches.

\noindent\textbf{Limitations.}
Our method assumes Gaussians are well aligned to scene surfaces, so it can be more effective when surface geometry is accurate.
In future work, we will further improve surface alignment by integrating geometrically faithful approaches~\cite{2dgs} and regularizers such as distortion losses, which would enhance both the stability of Top-K selection and overall reconstruction quality.


{
\bibliographystyle{splncs04}
\bibliography{main}
}
\end{document}